\documentclass[conference]{IEEEtran}
\IEEEoverridecommandlockouts
\usepackage{cite}
\usepackage{amsmath,amssymb,amsfonts}
\usepackage{algorithmic}
\usepackage{graphicx}
\usepackage{textcomp}
\usepackage{subcaption}
\usepackage{stfloats}
\usepackage{tcolorbox}
\usepackage{makecell}
\tcbuselibrary{skins,breakable}
\usepackage{amsthm}
\usepackage{xcolor}
\usepackage{hyperref}
\newtheorem{theorem}{Theorem}
\def\BibTeX{{\rm B\kern-.05em{\sc i\kern-.025em b}\kern-.08em
    T\kern-.1667em\lower.7ex\hbox{E}\kern-.125emX}}
\begin{document}

\title{Beyond Discrete Ancestry Groups: Distribution-Aware In-Context Learning for Cancer Outcome Prediction Across the Genetic Ancestry Continuum\\
\thanks{\textsuperscript{*}Corresponding authors: Weizhi Li (wli66@uthsc.edu) and Yan Cui (ycui2@uthsc.edu).}
}


\author{
\IEEEauthorblockN{1\textsuperscript{st} Weizhi Li\textsuperscript{*}}
\IEEEauthorblockA{
\textit{Department of Genetics, Genomics and Informatics} \\
\textit{University of Tennessee Health Science Center}\\
Memphis, TN, USA \\
wli66@uthsc.edu
}
\and
\IEEEauthorblockN{2\textsuperscript{nd} Yan Cui\textsuperscript{*}}
\IEEEauthorblockA{
\textit{Department of Genetics, Genomics and Informatics} \\
\textit{University of Tennessee Health Science Center}\\
Memphis, TN, USA \\
ycui2@uthsc.edu
}
}
\maketitle

\begin{abstract}
Machine learning models for omics-based prediction of cancer outcomes are commonly built on aggregated data without distinguishing ancestry groups, or under an ancestry-discrete design, which partitions samples into ancestry groups and fits a separate model within each group. Recent evidence shows that this approach is inherently limited: predictive accuracy declines with genetic distance from the training samples, even within a single ancestry group, because the data distribution shifts continuously rather than only between predefined categories. In this paper, we propose distribution-aware in-context learning (DA-ICL) for ancestry-continuum modeling. Our method extends conventional ancestry-discrete modeling to ancestry-continuum modeling, capturing the nonstationary omics-endpoint relationship across the genetic continuum while enabling efficient modeling through an in-context learning foundation model. We provide theoretical analysis demonstrating the superiority of the ancestry-continuum over ancestry-discrete modeling, along with empirical results supporting the proposed framework. Source code is available at \url{https://github.com/ai4pm/DA-ICL-Continuum}.
\end{abstract}


\section{Introduction}
\begin{figure}[t]
    \centering

    \begin{minipage}{1\linewidth}
        \centering
        \includegraphics[width=\linewidth]{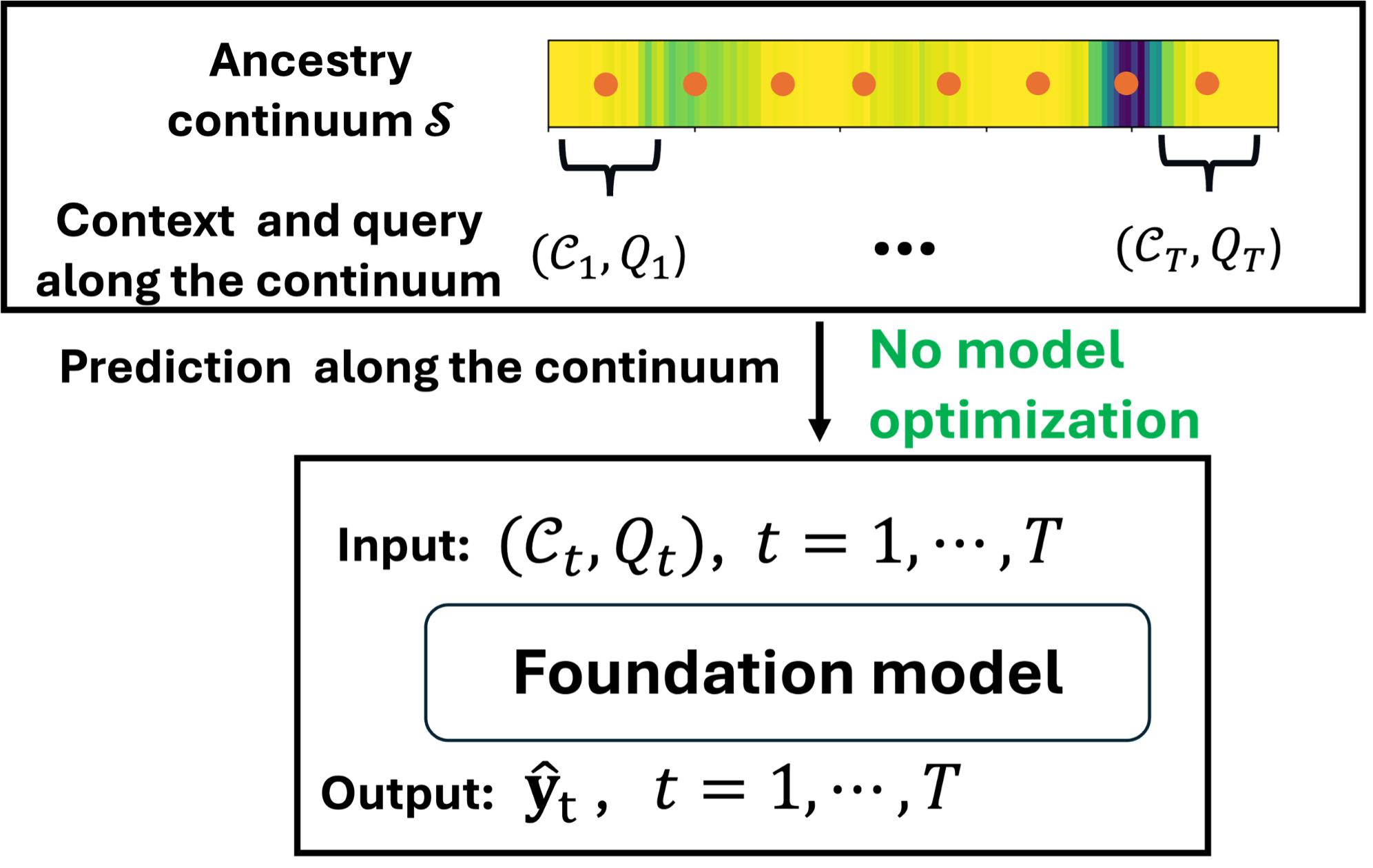}
        \textbf{(a) ICL for ancestry-continuum modeling}
    \end{minipage}

    \vspace{1em}

    \begin{minipage}{1\linewidth}
        \centering
        \includegraphics[width=\linewidth]{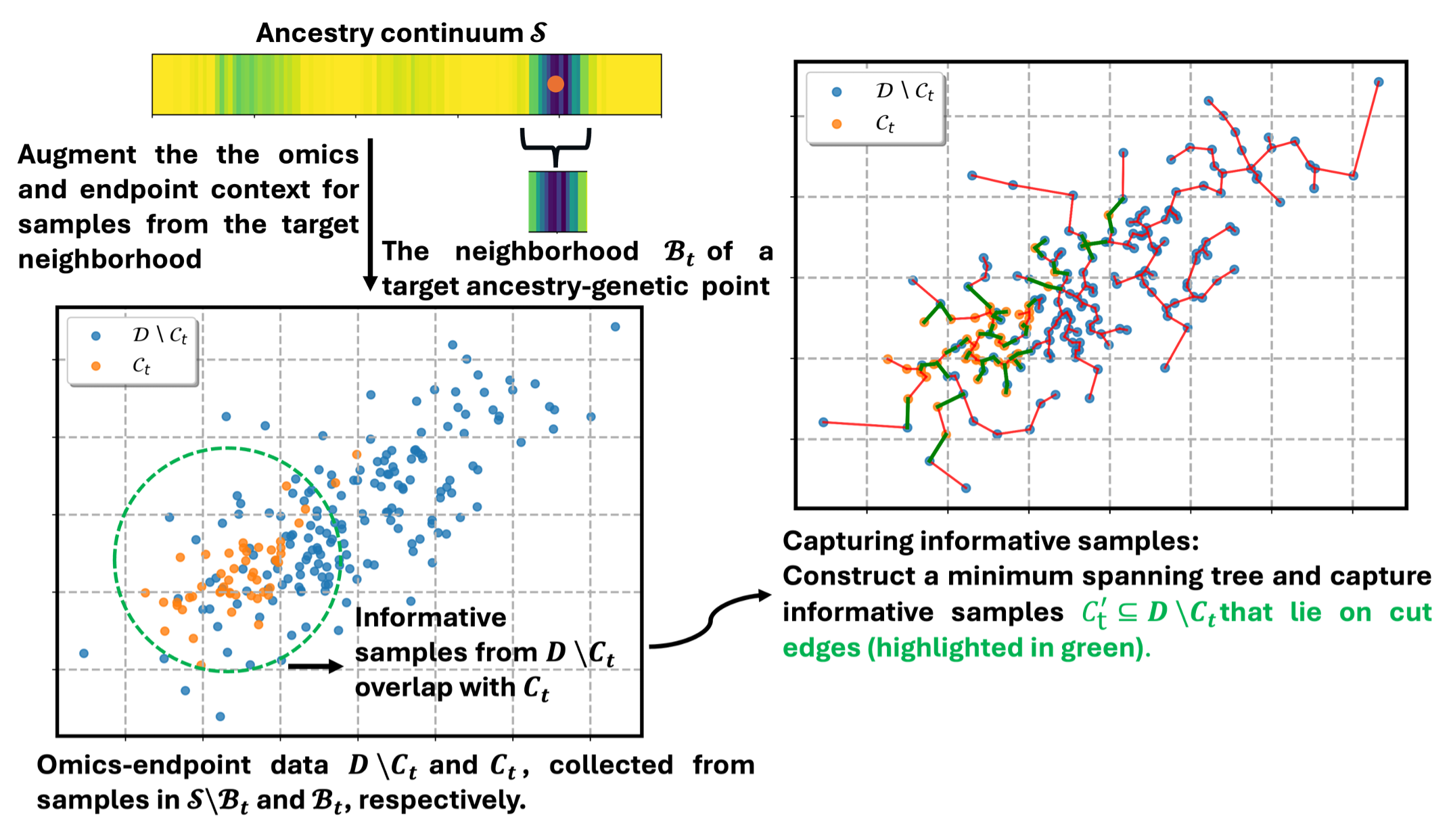}

        \textbf{(b) Distribution-aware context enhancement}
    \end{minipage}

    \caption{Distribution-aware ICL for the ancestry-continuum modeling.}
    \label{fig:illustration}
\end{figure}

Machine learning models are increasingly used to predict cancer-related risks and clinical outcomes from omics data. However, modeling without incorporating ancestral background often results in performance disparities across ancestry groups. This is fundamentally attributable to shifts in omics-feature distributions and in the relationships between omics features and phenotypes or clinical endpoints across populations with different ancestral backgrounds. A line of existing research~\cite{reddy2024bridging, petty2025large, gao2020deep, gao2023addressing, martin2019clinical, zhao2022construction} has performed ancestry-discrete modeling, which divides samples into ancestry groups, such as African (AFR) and European (EUR) populations, and conducts group-based analysis or prediction using omics features and phenotype or clinical-endpoint labels within each ancestry group. Such ancestry-discrete modeling has achieved predictive success, often outperforming global modeling approaches that do not incorporate ancestry information.

However, a recent study~\cite{ding2023polygenic} indicates that ancestry-discrete modeling is inherently limited. Using large biobanks, the authors of~\cite{ding2023polygenic} quantified genomic prediction performance at the individual level and showed that accuracy decreases approximately linearly with genetic distance from the training set, even within groups traditionally labeled as the same ancestry. Although a subsequent study~\cite{wang2026three} disagrees on how strongly genetic distance predicts accuracy, both studies identify the fundamental limitations of relying on broad, categorical ancestry labels in omics-based prediction of diseases and phenotypes.

In this paper, we propose distribution-aware in-context learning (DA-ICL) to address omics-based prediction along an ancestry continuum. To construct the ancestry continuum, we project genotype data into a low-dimensional space using principal component analysis (PCA). Recent studies~\cite{mcclelland2026multi, price2006principal, karczewski2024pan} have demonstrated that such low-dimensional genetic spaces can effectively capture ancestral disparities. Here we therefore perform omics-based modeling by incorporating continuous ancestry-genetic variables, aiming to capture the continuously shifting relationship between omics features and clinical-endpoint labels along the ancestry-genetic continuum.

Our proposed framework, illustrated in Figure~\ref{fig:illustration}, provides two natural components for ancestry-genetic continuum modeling. First, as shown in Figure~\ref{fig:illustration}(a), we apply a foundation model pretrained for in-context learning (ICL) to model the relationship between omics features and clinical endpoints at each ancestral point in the genetic continuum, using samples that are genetically close to that point. For each ancestral point, omics-endpoint data from local samples are used to construct a localized context. The ICL 
foundation model then adapts its prediction at inference time by conditioning on this context, enabling modeling along the genetic continuum \textbf{without target-task parameter updates to the foundation model}. Second, as shown in Figure~\ref{fig:illustration}(b), we propose distribution-aware context enhancement (DACE), which leverages informative samples across the entire ancestry-genetic continuum to augment the localized context for a target ancestral point. Specifically, DACE searches for samples that are similar to the nearby samples of the target ancestral point in terms of the distributional proximity of their omics features and endpoint labels, and incorporates them to construct a more informative context.

Our contributions can be summarized as follows:
\begin{itemize}
    \item We propose DA-ICL to enable efficient modeling along the ancestry-genetic continuum without target-task parameter updates to the foundation model.
    \item We provide theoretical analysis showing that the Bayes-optimal ancestry-continuum classifier has error no greater than the Bayes-optimal ancestry-discrete classifier under the stated data-generating process.
  \item We substantiate DA-ICL using TabPFN~\cite{hollmann2025accurate} and conduct experiments with synthetic data and omics data from The Cancer Genome Atlas Program (TCGA). The empirical evaluations demonstrate task-dependent benefits of ancestry-genetic continuum modeling and DACE.
\end{itemize}
 
\section{Related Work}

Many large-scale omics studies, including TCGA, TARGET, and GTEx, which are available through the National Cancer Institute’s Genomic Data Commons (GDC)~\cite{heath2021nci}, exhibit substantial imbalance in ancestral representation. Individuals of European (EUR) ancestry comprise the dominant proportion of samples, e.g., 83.3\% in TCGA, 69.7\% in TARGET, and 85\% in GTEx. Consequently, most existing machine-learning approaches for equitable omics-based prediction have been developed under a discrete-ancestry design.

A wide range of modeling approaches has been explored to mitigate ancestry bias in omics-based prediction under the ancestry-discrete modeling setting. Conventional tabular ML methods, such as elastic net~\cite{zou2005regularization}, XGBoost~\cite{chen2016xgboost}, and random forest~\cite{ho1995random}, are typically trained using samples from the target ancestry group to make predictions for that population. Transfer learning (TL) further leverages information from data-rich populations to improve prediction for underrepresented groups: a model is first pretrained using samples from data-rich ancestry groups and then fine-tuned on samples from the target data-disadvantaged ancestry group. Recent studies~\cite{sharma2025omics, gao2020deep, gao2024optimizing} have extensively explored TL pipelines in which machine-learning models, such as neural networks, are pretrained on data-rich ancestry data and subsequently fine-tuned for data-disadvantaged ancestry groups, showing promising results in improving predictive performance for underrepresented populations~\cite{gao2024optimizing, sharma2025omics, gao2020deep}.

Recently, the Tabular Prior-Fitted Network (TabPFN)~\cite{hollmann2025accurate}, a pretrained tabular foundation model, has demonstrated strong predictive performance and scalability relative to traditional tabular learning methods. TabPFN is trained on a large number of synthetic tasks using a transformer architecture~\cite{vaswani2017attention}, and it enables in-context learning by adapting predictions to the provided context at inference time. Several extensions~\cite{qu2025tabicl, zeng2025tabflex, kolberg2025tabpfn} have since been proposed, including TabPFN-Wide~\cite{kolberg2025tabpfn}, which has shown effectiveness in cancer risk prediction tasks using TCGA data. These developments highlight a shift from traditional model-optimization-based tabular methods toward in-context-learning-based foundation models for omics-based prediction tasks.

Recent work~\cite{ding2023polygenic, ruan2025discodivas, chen2024splendid, wang2026three} has discussed the inherent limitations of ancestry-discrete modeling, highlighting the need for ancestry-continuum modeling to advance precision medicine. However, existing approaches either originate from ancestry-discrete models, rather than directly characterizing the omics--clinical endpoint relationship for samples along the ancestry continuum, or rely on oversimplified statistical modeling assumptions for omics features and endpoints. In this paper, we propose DA-ICL, which incorporates a state-of-the-art tabular foundation model to directly model omics-endpoint relationships for each ancestry point along the continuum.
\section{Data-generating process}
\label{sec: dgp}
We write $\left(\mathbf{X}, \mathbf{Y}\right)$ to denote a pair consisting of omics features and clinical-endpoint labels, and write $p\left(\mathbf{x}, \mathbf{y}\right)$ to denote the probability density function (pdf) from which a set of \textit{i.i.d.} pairs $\mathcal{D}=\left\{\left(\mathbf{X}_i, \mathbf{Y}_i\right)\right\}_{i=1}^n$ are generated. In addition, we use $\mathbf{S}$ to denote genetic ancestry variables, representing genetic features that characterize ancestral disparities. Similarly, we write $p\left(\mathbf{s}\right)$ to denote the pdf from which the \textit{i.i.d.} sequence $\left\{\mathbf{S}_i\right\}_{i=1}^n$ is generated. Lastly, we use $\mathcal{X}$ and $\mathcal{S}$ to denote the omics-feature space and the ancestry-genetic space that support $p\left(\mathbf{x}\right)$ and $p\left(\mathbf{s}\right)$, respectively.

One can condition the omics-endpoint variables $\left(\mathbf{X}, \mathbf{Y}\right)$ on the ancestry-genetic variable $\mathbf{S}$, resulting in $p\left(\mathbf{x}, \mathbf{y}\mid \mathbf{s}\right)$, which indicates that the joint pdf of the omics-endpoint variables varies along $\mathbf{s}$. In a large body of literature (e.g.,~\cite{sharma2025omics, gao2024optimizing, huang2026admixture, gao2021multi}), the ancestry-genetic space $\mathcal{S}$  is divided into $C$ disjoint partitions $\{\mathcal{P}_1,\cdots, \mathcal{P}_C\}$, such as European, African, and Asian ancestry groups. Using $c$ to denote the ancestry assignment, $p\left(\mathbf{x}, \mathbf{y}\mid c\right), 1\leq c\leq C$ then indicates that the distribution of the omics-endpoint variables varies across ancestry group $c$.

By the rules of probability, we have
\begin{align}
p\left(\mathbf{x},\mathbf{y}\right)
&= \int_{\mathcal{S}}p\left(\mathbf{x},\mathbf{y}\mid \mathbf{s}\right)p(\mathbf{s})d\mathbf{s}\nonumber\\ 
&= \sum_{c=1}^C p\left(\mathbf{x},\mathbf{y}\mid c\right)P\left(\mathbf{S}\in \mathcal{P}_c\right),
\label{eq:dgp}
\end{align}
which indicates the relationship between the joint pdf of $\left(\mathbf{X}, \mathbf{Y}\right)$ and its conditional pdfs given the ancestry-genetic variable $\mathbf{s}$ and ancestry group variable $c$, respectively.

\section{Superiority of Ancestry-Continuum Modeling over Ancestry-Discrete Modeling}

\subsection{Ancestry-Continuum vs. Ancestry-Discrete Probabilistic Modeling}
\label{sec: probalistic_modeling}

\textbf{Ancestry-discrete probabilistic modeling} divides the observed triples of omics features, clinical-endpoint labels, and ancestry-genetic variables, $\mathcal{D}=\left\{\left(\mathbf{x}_i, \mathbf{y}_i, \mathbf{s}_i\right)\right\}_{i=1}^n$, into $C$ disjoint ancestry groups $\left\{\mathcal{D}_c\right\}_{c=1}^C$, according to the genetic distance between $\mathbf{s}_i$ and the centroid of $\mathcal{P}_c$ for $1\leq c\leq C$, where $\mathcal{D}=\bigcup_{c=1}^C\mathcal{D}_c$. The omics-based prediction task is then performed separately for each ancestry group using $\mathcal{D}_c$, resulting in models that approximate $p\left(\mathbf{y}\mid \mathbf{x},c\right)$ for $1\leq c\leq C$. As a result, the prediction for an omics feature vector $\mathbf{x}$ is made using $p\left(\mathbf{y}\mid \mathbf{x}, c\right)$, depending on the ancestry group $c$ to which the corresponding ancestry-genetic variable $\mathbf{s}$ is assigned. Therefore, ancestry-discrete modeling makes group-specific but within-group homogeneous predictions for samples in the same ancestry partition.

\textbf{Ancestry-continuum probabilistic modeling} incorporates the continuous ancestry-genetic variable $\mathbf{s}$ into the modeling, rather than stratifying $\left(\mathbf{x}, \mathbf{y}\right)$ by discrete ancestry groups. It models the relationship between the label variable $\mathbf{y}$ and the feature variable $\mathbf{x}$ across the ancestry-genetic space $\mathcal{S}$, and therefore approximates $p\left(\mathbf{y}\mid \mathbf{x}, \mathbf{s}\right)$. Ancestry-continuum modeling makes predictions adapted to the ancestry-genetic variable $\mathbf{s}$ rather than to a discrete ancestry group $c$.

A study~\cite{ramos2026polygenic} has illustrated the empirical performance degradation of ancestry-discrete modeling for samples whose genetic features $\mathbf{s}$ are distant from the centroids of the ancestry partitions over $\mathcal{S}$. In Section~\ref{sec:bayes_error_rates}, we provide a theoretical analysis characterizing the superiority of ancestry-continuum modeling over ancestry-discrete modeling.

\subsection{Superiority of Optimal Ancestry-Continuum Modeling over Ancestry-Discrete Modeling}
\label{sec:bayes_error_rates}

Ancestry-continuum and ancestry-discrete modeling make predictions given $\left(\mathbf{x}, \mathbf{s}\right)$ and $\left(\mathbf{x}, c\right)$, respectively. Under the data-generating process (DGP) defined in Section~\ref{sec: dgp}, this leads to the optimal classifiers
\begin{align*}
\hat{\mathbf{y}}_{\text{cont}}
&= \arg\max_{\mathbf{y}} p\left(\mathbf{y} \mid \mathbf{x}, \mathbf{s}\right),\\
\hat{\mathbf{y}}_{\text{dis}}
&= \arg\max_{\mathbf{y}} p\left(\mathbf{y} \mid \mathbf{x}, c\right),
\end{align*}
for ancestry-continuum and ancestry-discrete modeling. Consequently, we have the following theorem.
\begin{theorem}
\label{theorem:superiority}
Define a classification problem with joint distribution $\left(\mathbf{X}, \mathbf{Y}, \mathbf{S}\right)\sim p\left(\mathbf{x}, \mathbf{y}, \mathbf{s}\right)$, and let $\left\{\mathcal{P}_1,\cdots, \mathcal{P}_C\right\}$ be $C$ disjoint ancestry partitions such that $\mathcal{S}=\bigcup_{c=1}^C\mathcal{P}_c$. Then, the optimal solutions of ancestry-continuum and ancestry-discrete modeling achieve error rates $\epsilon^*_{\text{cont}}$ and $\epsilon^*_{\text{dis}}$, respectively, where
\begin{align}
\epsilon_{\text{cont}}^*
&= \int_\mathcal{S} \int_{\mathcal{X}}
\left(1 - p\left(\hat{\mathbf{y}}_{\text{cont}} \mid \mathbf{x}, \mathbf{s}\right)\right)
p\left(\mathbf{x} \mid \mathbf{s}\right)p\left(\mathbf{s}\right)
d\mathbf{x}d\mathbf{s},
\label{eq:bayes_cont}\\
\epsilon_{\text{dis}}^*
&= \sum_{c=1}^C P\left(\mathbf{S}\in \mathcal{P}_c\right)
\int_\mathcal{X} 
\left(1 - p\left(\hat{\mathbf{y}}_{\text{dis}} \mid \mathbf{x}, c\right)\right)
p\left(\mathbf{x} \mid c\right)
d\mathbf{x}.
\label{eq:bayes_dis}
\end{align}
In addition, given the Bayes error rate $\epsilon^*$, which denotes the lowest achievable error rate for the considered classification problem, we have
$
\epsilon^*=\epsilon^*_{\text{cont}}\leq \epsilon^*_{\text{dis}}.
$
\end{theorem}

\begin{proof}
Let $\mathbb{C}(\mathbf{s})$ denote the ancestry-group assignment of $\mathbf{s}$, i.e., $\mathbb{C}(\mathbf{s})=c$ if $\mathbf{s}\in\mathcal{P}_c$. The ancestry-discrete optimal classifier can then be written as $\hat{\mathbf{y}}_{\text{dis}}\left(\mathbf{x},\mathbb{C}(\mathbf{s})\right)$. Rewriting~\eqref{eq:bayes_dis} over the continuous ancestry space $\mathcal{S}$ gives
$
\epsilon^*_{\text{dis}}
=
\int_{\mathcal{S}}\int_{\mathcal{X}}
\left(
1-
p\left(
\hat{\mathbf{y}}_{\text{dis}}\left(\mathbf{x},\mathbb{C}(\mathbf{s})\right)
\mid \mathbf{x},\mathbf{s}
\right)
\right)
p\left(\mathbf{x}\mid \mathbf{s}\right)p\left(\mathbf{s}\right)
d\mathbf{x}d\mathbf{s}.
$
Therefore,
\begin{align}
\epsilon^*_{\text{dis}}-\epsilon^*_{\text{cont}}
&=
\int_{\mathcal{S}}\int_{\mathcal{X}}
\left[
p\left(\hat{\mathbf{y}}_{\text{cont}} \mid \mathbf{x},\mathbf{s}\right)
-
p\left(
\hat{\mathbf{y}}_{\text{dis}}\left(\mathbf{x},\mathbb{C}(\mathbf{s})\right)
\mid \mathbf{x},\mathbf{s}
\right)
\right] \nonumber \\
&\qquad \times
p\left(\mathbf{x}\mid \mathbf{s}\right)p\left(\mathbf{s}\right)
d\mathbf{x}d\mathbf{s}.
\end{align}

Since $\hat{\mathbf{y}}_{\text{cont}}$ is the Bayes-optimal classification label given $\left(\mathbf{x},\mathbf{s}\right)$, we have
$
p\left(\hat{\mathbf{y}}_{\text{cont}} \mid \mathbf{x},\mathbf{s}\right)
\geq
p\left(
\hat{\mathbf{y}}_{\text{dis}}\left(\mathbf{x},\mathbb{C}(\mathbf{s})\right)
\mid \mathbf{x},\mathbf{s}
\right)
$
for all $\left(\mathbf{x},\mathbf{s}\right)$. Hence,
$
\epsilon^*=\epsilon^*_{\text{cont}}\leq \epsilon^*_{\text{dis}}.
$
\end{proof}

Theorem~\ref{theorem:superiority} follows from the fact that the optimal ancestry-continuum classifier accounts for variation in $\mathbf{s}$, whereas ancestry-discrete modeling discards within-group ancestry heterogeneity. As a result, ancestry-discrete modeling makes predictions based only on the discrete ancestry group $c$, which is generally suboptimal when the omics-endpoint relationship varies along the ancestry-genetic space. When Bayes-optimal classifiers are selected under both the ancestry-continuum and ancestry-stratified settings, the ancestry-continuum classifier achieves an error rate no greater than that of the ancestry-stratified classifier. This comparison concerns optimal classifiers under the stated data-generating process and does not guarantee that a finite-sample implementation of DA-ICL will outperform an ancestry-stratified model.

\section{Distribution-Aware In-Context Learning for Ancestry-Continuum Modeling}

Figure~\ref{fig:illustration} illustrates our proposed method, distribution-aware in-context learning (DA-ICL), for ancestry-continuum modeling. Section~\ref{sec:ICL_continuum} presents ancestry-continuum modeling using ICL. Section~\ref{sec:DACE} presents distribution-aware context enhancement (DACE), which is proposed to address the sparse-context issue in ancestry-continuum modeling.

\subsection{Efficient Ancestry-Continuum Modeling via ICL}
\label{sec:ICL_continuum}

ICL is a capability of transformer-based foundation models, popularized by large language models, that enables predictions to be generated from a context provided at inference time without target-task parameter updates to the foundation model. Figure~\ref{fig:illustration}(a) illustrates that ICL efficiently models along the ancestry continuum by constructing omics and clinical-endpoint contexts $\mathcal{C}$ that vary along the ancestry-genetic space and by making predictions for each target ancestry point $\mathbf{s}\in\mathcal{S}$.

The authors of~\cite{price2006principal} demonstrate that projecting genotype data into a low-dimensional space $\mathcal{S}$ using principal component analysis (PCA) can capture ancestral disparities. Herein, \textbf{we posit that ancestry-continuum modeling is performed along the low-dimensional space $\mathcal{S}$ projected from the original genotype data using PCA}.

Let $\mathcal{D}=\{(\mathbf{x}_i,\mathbf{y}_i)\}_{i=1}^{n}$ denote the full omics-endpoint dataset, where $\mathbf{x}_i$ is the omics feature vector and $\mathbf{y}_i$ is the corresponding clinical-endpoint label. For a target ancestral point $\mathbf{s}_t\in\mathcal{S}$, let $\mathcal{B}_t\subseteq\mathcal{S}$ denote its neighborhood in the ancestry-genetic space $\mathcal{S}$. The ICL context set is then constructed as the subset of samples whose ancestry points lie in $\mathcal{B}_t$:
$
\mathcal{C}_t
=
\{(\mathbf{x}_i,y_i)\in \mathcal{D}: \mathbf{s}_i\in\mathcal{B}_t\},
$
where $|\mathcal{C}_t|=n_t$. In practice, $\mathcal{C}_t$ can be developed using a $k$-nearest-neighbor (KNN) search based on the proximity to $\mathbf{s}_t$ in the space of $\mathcal{S}$.  The omics feature vector $\mathbf{x}_{t}$ associated with the target ancestral point $\mathbf{s}_{t}$ is treated as the query input $\mathcal{Q}_t$.

The context set $\mathcal{C}_t$ represents omics-endpoint pairs and, depending on the biological setting, may include omics features obtained from non-inherited molecular profiling, such as transcriptomics, or inherited omics, such as significant SNPs identified from GWAS, along with disease phenotypes or clinical outcomes. The omics features provided for $\mathcal{Q}_t$ at the target ancestral point are of the same type as those included in $\mathcal{C}_t$. Given $(\mathcal{C}_t,\mathcal{Q}_t)$, a pretrained foundation model, such as TabPFN~\cite{hollmann2025accurate}, predicts the corresponding query label $\hat{\mathbf{y}}_{t}$, e.g., cancer status or clinical outcome, without target-task parameter updates. $\hat{\mathbf{y}}_{t}$ is then treated as the classification prediction for the target omics-ancestry pair $\left(\mathbf{x}_{t}, \mathbf{s}_{t}\right)$.

\subsection{Distribution-Aware Context Enhancement (DACE)}
\label{sec:DACE}

This section introduces an approach to enhance the context $\mathcal{C}_t$ for a target ancestry point $\mathbf{s}_t$ by leveraging informative examples from $\mathcal{D}\setminus \mathcal{C}_t$.  $\mathcal{D}\setminus \mathcal{C}_t$ consists of omics-feature and clinical-endpoint pairs that are not included in $\mathcal{C}_t$ because their corresponding genetic ancestry points are distant from $\mathbf{s}_t$.

Our approach is inspired by the nonparametric $f$-divergence estimation method proposed in~\cite{berisha2015empirically} and subsequently extended to a range of applications in~\cite{li2026regularization, li2024active, li2024advanced, li2022label}. We use $p_{t}\left(\mathbf{x}, \mathbf{y}\mid \mathcal{B}_t\right)$ and $p_{\text{all}\setminus t}\left(\mathbf{x}, \mathbf{y}\mid \mathcal{S}\setminus\mathcal{B}_t\right)$ to denote the pdfs from which samples in $\mathcal{C}_t$ and $\mathcal{D}\setminus \mathcal{C}_t$ are \textit{i.i.d.} generated, respectively. Figure~\ref{fig:illustration}(b) illustrates DACE, which leverages samples from $\mathcal{D}\setminus\mathcal{C}_t$ that are close to $\mathcal{C}_t$ in terms of distributional proximity, measured by $f$-divergence.

\paragraph{$f$-divergence estimation}
Theorem~1 in~\cite{berisha2015empirically} states that the $f$-divergence  first proposed in~\cite{csiszar1963informationstheoretische} to characterize the distance between two pdfs, e.g., $p_t\left(\mathbf{x}, \mathbf{y}\mid \mathcal{B}_t\right)$ and $p_{\text{all}\setminus t}\left(\mathbf{x}, \mathbf{y}\mid \mathcal{S}\setminus\mathcal{B}_t\right)$, can be estimated by first constructing a minimum spanning tree (MST) over their two realization sets, $\mathcal{C}_t$ and $\mathcal{D}\setminus \mathcal{C}_t$, and then counting the number of cut edges that connect points across the two datasets. The proportion of such cut edges among the total number of edges provides an approximation of $1 - D_f\left(p_t \,\|\, p_{\text{all}\setminus t}\right)$, where $D_f$ denotes the $f$-divergence between two pdfs. Under this setting,
\[
\lim_{n \to \infty}
\left(
1 - \#\text{cut-edge} \times \frac{n}{2 n_t n_{\text{all}\setminus t}}
\right)
=
D_f\left(p_t \,\|\, p_{\text{all}\setminus t} \right),
\]
as demonstrated in~\cite{berisha2015empirically} for a chosen $f$-function, where $n = |\mathcal{D}|$, $n_t = |\mathcal{C}_t|$, and $n_{\text{all}\setminus t} = |\mathcal{D}\setminus \mathcal{C}_{t}|$.

\paragraph{Distribution-aware context enhancement (DACE)}
Box~\ref{box:CE} illustrates DACE, which constructs a subset $\mathcal{C}_t' \subseteq \mathcal{D} \setminus \mathcal{C}_t$ that is distributionally close to $\mathcal{C}_t$. The samples in $\mathcal{C}_t'$ are then combined with the original context $\mathcal{C}_t$, due to their distributional similarity, to enrich the context for ICL.  This design is motivated by $D_f$ estimation in Section~\ref{sec:DACE}(a), where samples lying on cut edges connect $\mathcal{C}_t$ and $\mathcal{D}\setminus \mathcal{C}_t$ in regions of high distributional similarity and thus contribute most to reducing the divergence $D_f\left(p_t \,\|\, p_{\text{all}\setminus t}\right)$.

%
\newtcolorbox[auto counter]{mytcolorbox}[2][]{%
  enhanced,
  breakable,
  colback=white,
  colframe=black,
  fonttitle=\bfseries,
  title={Box~\thetcbcounter: #2},
  #1
}

\begin{mytcolorbox}[label={box:CE}]{Distribution-Aware Context Enhancement}
Let $\mathcal{D}= \{(\mathbf{x}_i, \mathbf{y}_i)\}_{i=1}^n$ denote the omics features and clinical-endpoint labels for all samples, and let $\mathcal{C}_t \subseteq \mathcal{D}$ denote the subset corresponding to a target ancestry point $\mathbf{s}_t\in\mathcal{S}$. Construct a minimum spanning tree $G = (V, E)$ over $\mathcal{D}$. We then construct
$
\mathcal{C}_t'
=
\left\{
v_0 \in \mathcal{D}\setminus\mathcal{C}_t
\;\middle|\;
\substack{\exists v_1 \in \mathcal{C}_t\\
(v_0, v_1) \in E}
\right\},
$
and use $\mathcal{C}_t \cup \mathcal{C}_t'$ as the context for ICL at the target ancestry point $\mathbf{s}_t\in\mathcal{S}$.
\end{mytcolorbox}
\section{Experimental results}
We perform experiments using synthetic data and BRCA, KIRC, and UCEC datasets from The Cancer Genome Atlas Program (TCGA)~\cite{weinstein2013cancer}. These experiments evaluate two questions: (1) whether modeling over the ancestry-continuum improves predictive performance compared with ancestry-discrete modeling, and (2) whether incorporating DACE into ICL further enhances modeling over the ancestry-continuum.

\subsection{Data description}
\label{sec:exp_dgp}
\paragraph{DGP for synthetic data}
The ancestry-genetic variable is a two-dimensional coordinate
$\mathbf{S} = \left(S_0, S_1\right) \in \left[0,1\right]^{2}$,
generated as a mixture of \emph{clustered} and \emph{admixed}
ancestry samples.

\textit{Clustered samples (65\%):}
A cluster label is drawn uniformly at random,
$c \sim \text{Uniform}\left\{1, 2, 3\right\}$,
and the ancestry-genetic coordinate is placed at the corresponding
centroid with isotropic Gaussian noise:
$
  \mathbf{S} = \boldsymbol{\mu}_{c} + \boldsymbol{\epsilon},
  \quad
  \boldsymbol{\epsilon} \sim \mathcal{N}\!\left(\mathbf{0},\, 0.055^{2}\,\mathbf{I}_{2}\right),
$
where the three ancestry centroids are
$\boldsymbol{\mu}_{1} = \left(0.15,\, 0.20\right)$,
$\boldsymbol{\mu}_{2} = \left(0.85,\, 0.25\right)$, and
$\boldsymbol{\mu}_{3} = \left(0.50,\, 0.85\right)$.

\textit{Admixed samples (35\%):}
An anchor pair is chosen uniformly from the set of centroid pairs,
$\left(\boldsymbol{\mu}_{i},\, \boldsymbol{\mu}_{j}\right)
  \sim \left\{
    \left(\boldsymbol{\mu}_{1}, \boldsymbol{\mu}_{2}\right),\;
    \left(\boldsymbol{\mu}_{1}, \boldsymbol{\mu}_{3}\right),\;
    \left(\boldsymbol{\mu}_{2}, \boldsymbol{\mu}_{3}\right)
  \right\}$,
and an interpolation weight is sampled as
$T \sim \text{Beta}\left(1.4,\, 1.4\right)$.
The ancestry-genetic coordinate is then
$
  \mathbf{S}
  = \left(1 - T\right)\,\boldsymbol{\mu}_{i}
  + T\,\boldsymbol{\mu}_{j}
  + \boldsymbol{\epsilon},
  \quad
  \boldsymbol{\epsilon} \sim \mathcal{N}\!\left(\mathbf{0},\, 0.035^{2}\,\mathbf{I}_{2}\right).
$
All coordinates are clipped to the unit square.
\textbf{The resulting ancestry-genetic distribution thus consists of three
noisy clusters connected by admixed samples along the lines joining
each pair of clusters}.

The omics feature vector
$\mathbf{X} = \left(X_0,\, X_1\right)$
is generated independently of the ancestry-genetic variable $\mathbf{S}$ as
$\mathbf{X} \sim \text{Uniform}\!\left(\left[-1,1\right]^{2}\right)$.
Thus, ancestry does not affect the marginal distribution of
$\mathbf{X}$; instead, it determines the decision boundary used to
generate a label variable $Y\in\{0, 1\}$ .

The baseline decision boundary is the diagonal $X_1 = X_0$,
whose unit normal vector is
$\mathbf{n}_{0} = \left(-1,\, 1\right)/\sqrt{2}$.
The first ancestry coordinate $S_0$ controls the rotation of this
boundary,
$\theta\!\left(S_0\right) = r\left(S_0 - 0.5\right)$,
where $r$ is the rotation-degree parameter, and
the second coordinate $S_1$ controls the boundary offset,
$b\!\left(S_1\right) = o\left(S_1 - 0.5\right)$,
where $o$ is the offset-degree parameter.
Let 
$\mathbf{n}\!\left(S_0\right)
  = \mathbf{R}_{\theta(S_0)}\,\mathbf{n}_{0}$
be the rotated normal vector, where $\mathbf{R}_{\theta}=\begin{bmatrix}
\cos{\theta} & -\sin{\theta} \\
\sin{\theta} & \cos{\theta} 
\end{bmatrix}$ denotes
the $2\!\times\!2$ rotation matrix.
The signed margin is then
$
  m\!\left(\mathbf{X},\, \mathbf{S}\right)
  = \mathbf{X}^{\top}\mathbf{n}\!\left(S_0\right)
    - b\!\left(S_1\right).
$
The clinical-endpoint label is sampled as
$
  Y \sim \text{Bernoulli}\!\left(
    \sigma\!\left(8\,m\!\left(\mathbf{X},\, \mathbf{S}\right)\right)
  \right),
$
where
$\sigma(z)$ represents a sigmoid function.
In summary, the ancestry-genetic variables $S_0$ and $S_1$
determine how the diagonal decision boundary is rotated and
shifted, respectively, before the binary label is sampled.
\textbf{Additionally, the parameters $r$ and $o$ control the rotation
and shifting degree for $S_0$ and $S_1$ on the baseline decision
boundary, and $r = 0$ and $o = 0$ indicates no impact of the
ancestry-genetic variables on the label-generating process}.

\paragraph{BRCA, KIRC and UCEC from TCGA}
We use three cancer types from TCGA: breast invasive carcinoma (BRCA, $n \approx 900$),
kidney renal clear cell carcinoma (KIRC, $n \approx 380$), and
uterine corpus endometrial carcinoma (UCEC, $n \approx 408$).
For each cancer type, three omics modalities---mRNA expression,
microRNA expression, and DNA methylation---are paired with two
clinical endpoints: overall survival (OS)
and progression-free interval (PFI),
yielding classification tasks to predict whether OS and PFI are
shorter than 2 years. A positive label denotes an observed OS or PFI
event before day 730. A negative label requires the patient to be
observed event-free through at least day 730; patients censored before
day 730 and patients with missing required data are excluded. This
results in 18 TCGA tasks. The eligible positive/total counts for
(mRNA, miRNA, methylation) were BRCA OS: (397/965, 391/938,
347/881), BRCA PFI: (397/947, 392/922, 345/860); KIRC OS:
(82/405, 78/377, 81/404), KIRC PFI: (102/377, 96/349, 100/376);
UCEC OS: (166/438, 165/432, 160/417), and UCEC PFI:
(162/401, 160/395, 156/382). We use omics PCA100 for prediction, genotype PCA30 to construct
the ancestry-genetic space~$\mathcal{S}$, and an omics autoencoder
embedding (AE20) for DACE. These unsupervised transformations were
fitted once on the available cohort-wide measurements without outcome
labels and reused across partitions. 
\begin{figure}[h]
\centering
\includegraphics[width=0.499\linewidth]{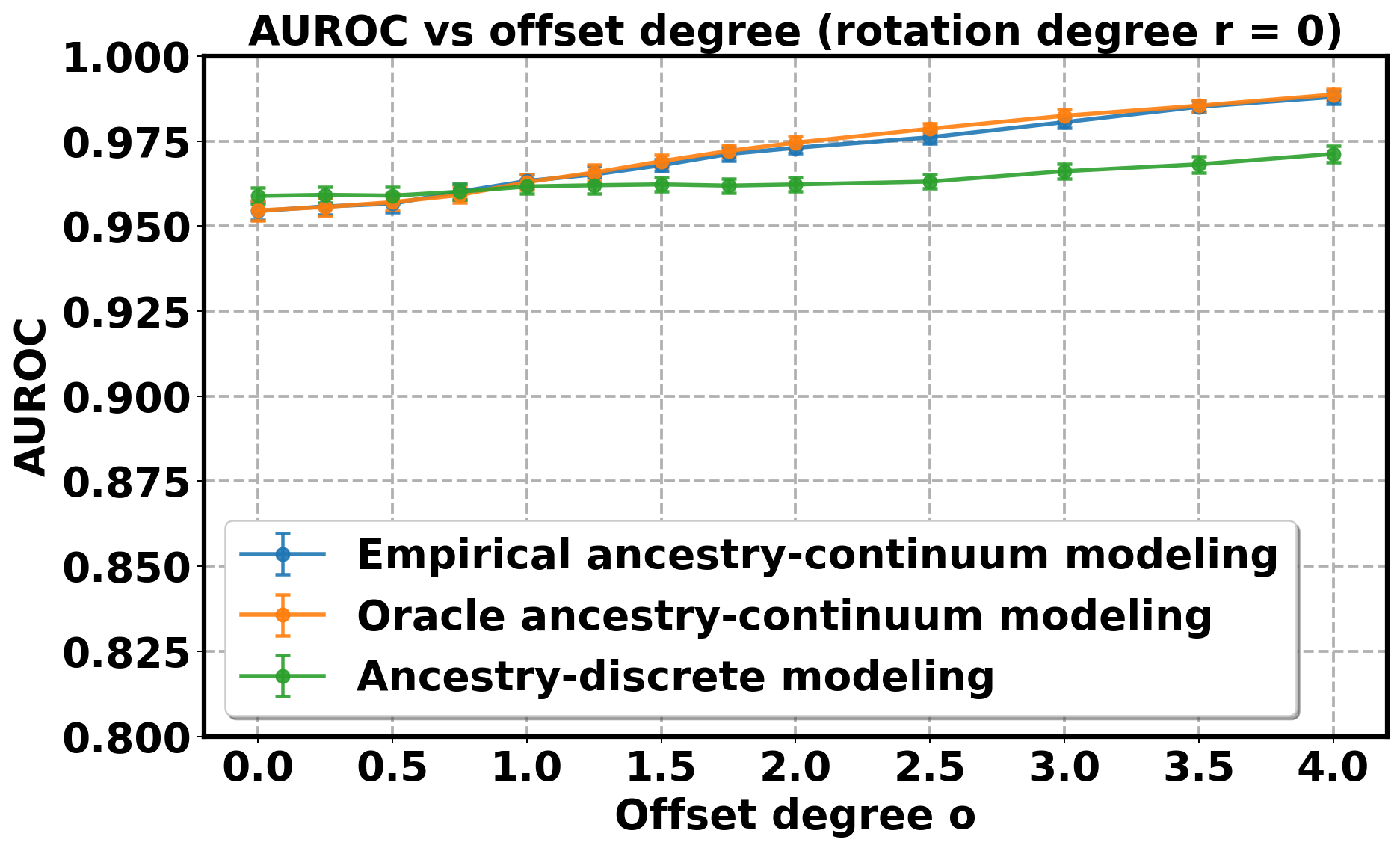}
\includegraphics[width=0.487\linewidth]{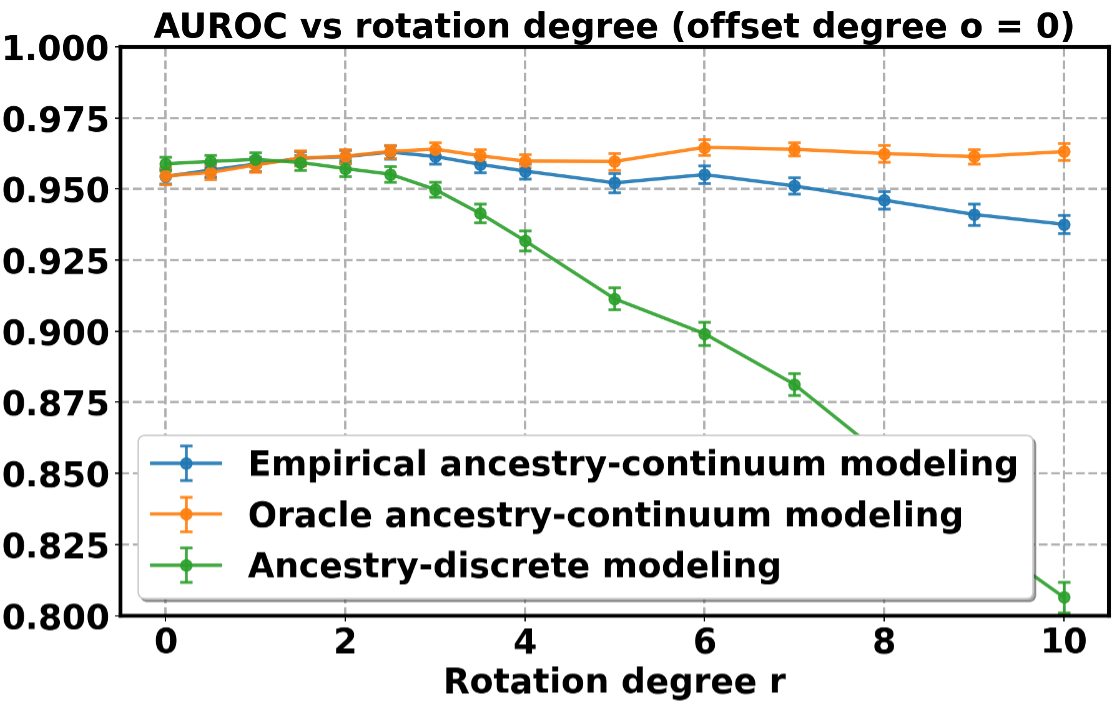}
\caption{AUROCs of ancestry-discrete, oracle ancestry-continuum, and empirical ancestry-continuum modeling on synthetic data with increasing offset degree~$o$ and rotation degree~$r$, respectively. Setting both $o$ and $r$ to 0 indicates that the relationship between $y$ and $\mathbf{x}$ does not vary across $\mathbf{s}$.}
\label{fig:syn_rotation_offset}
\end{figure}
\vspace{-0.5cm}
\begin{figure}[h]
  \centering
  \includegraphics[width=0.48\linewidth]{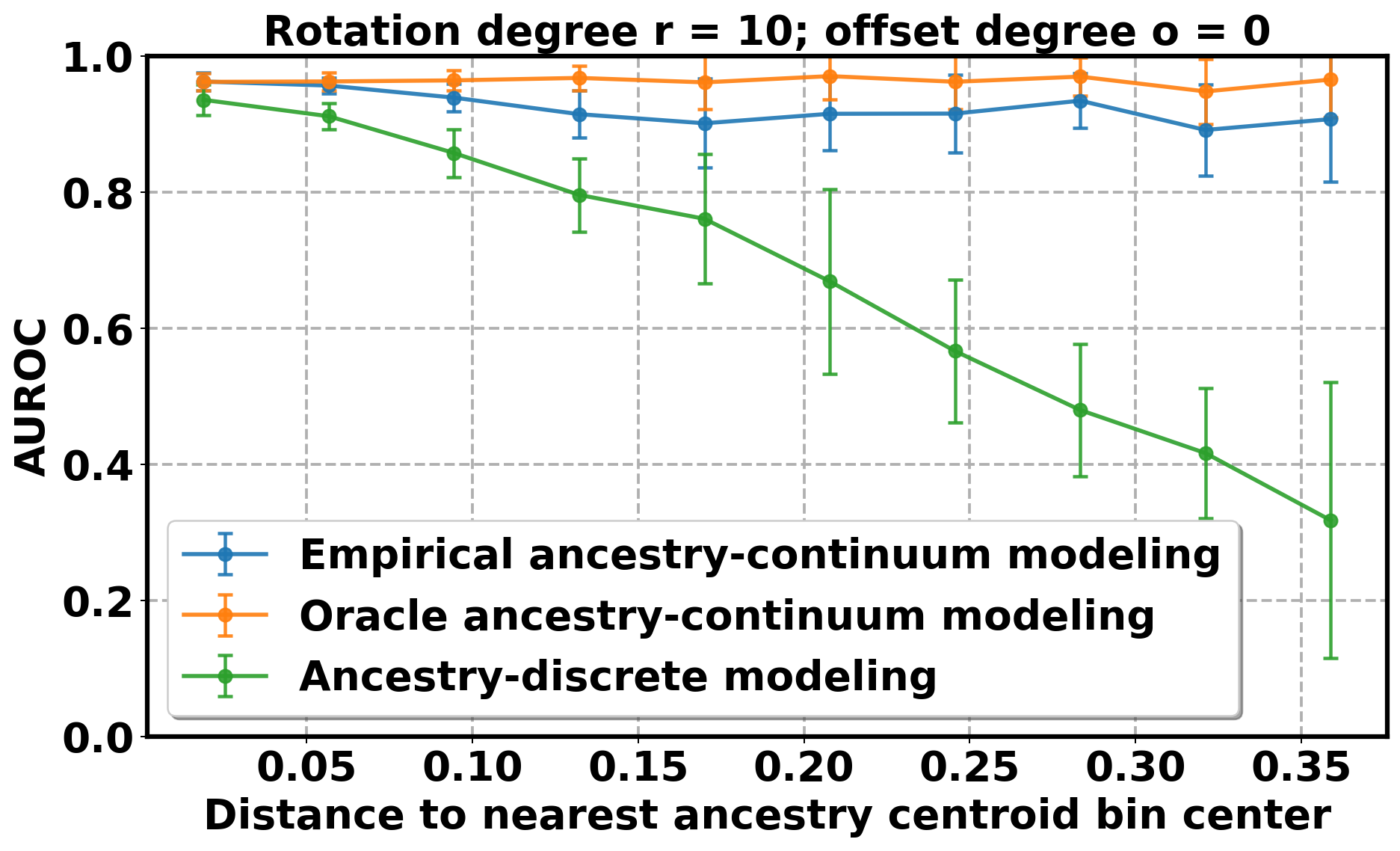}
  \includegraphics[width=0.48\linewidth]{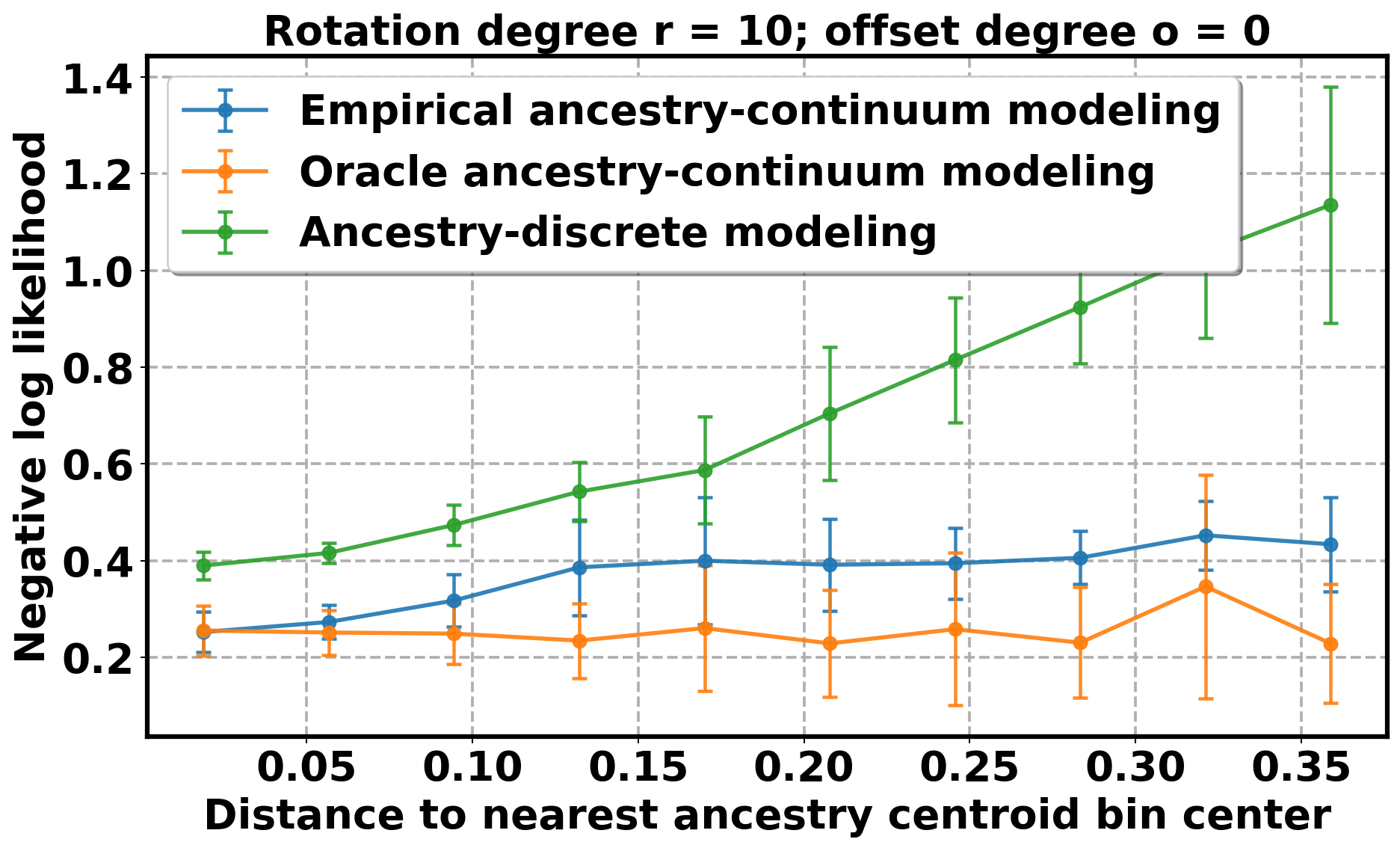}
\caption{Performance of ancestry-discrete, oracle ancestry-continuum, and empirical ancestry-continuum modeling on synthetic data as a function of the distance between the ancestry-genetic variable~$\mathbf{s}$ and its nearest ancestry-group centroid~$\boldsymbol{\mu} \in \{\boldsymbol{\mu}_1, \boldsymbol{\mu}_2, \boldsymbol{\mu}_3\}$. Higher AUROC and lower negative log-likelihood (NLL) indicate better performance.}
\label{fig:syn_distance}
\end{figure}
\subsection{Experiments with synthetic data}
\label{sec: exp_syn}
We generate 1{,}000 training and 600 test samples using the DGP
defined in Section~\ref{sec:exp_dgp}.
To empirically evaluate Theorem~\ref{theorem:superiority}, we
construct logistic regression models under three approaches:
(1)~\emph{\textbf{ancestry-discrete modeling}}, which partitions the
training set into $C$ disjoint ancestry groups
$\{\mathcal{P}_1, \dots, \mathcal{P}_C\}$ and fits a separate
model for each group (in the synthetic setting, $C = 3$);
(2)~\emph{\textbf{oracle ancestry-continuum modeling}}, which, for each
test sample with ancestry-genetic feature~$\mathbf{s}$,
repeatedly draws 100 samples $(\mathbf{X}, Y)$ from
$p(\mathbf{x}, y \mid \mathbf{s})$ and trains a model on them;
and
(3)~\emph{\textbf{empirical ancestry-continuum modeling}}, which, for each
test sample with ancestry-genetic feature~$\mathbf{s}$, trains a
model on the 100 training samples whose ancestry-genetic features
are closest to~$\mathbf{s}$. \textbf{The synthetic experiments do not evaluate DA-ICL, as their purpose is to independently assess the benefit of shifting from ancestry-discrete modeling to ancestry-continuum modeling, thereby supporting Theorem~\ref{theorem:superiority}.}

We perform the three modeling approaches under the same DGP 30 times. Figure~\ref{fig:syn_rotation_offset} presents the AUROC
as functions of the rotation-degree and offset-degree parameters,
$r$ and~$o$, which control the extent to which
$p(\mathbf{y} \mid \mathbf{x}, \mathbf{s})$ varies
over~$\mathbf{s}$.
Both \emph{oracle ancestry-continuum modeling} and \emph{empirical ancestry-continuum modeling} improve relative to ancestry-discrete modeling as $r$ and~$o$ increase, achieving substantially higher AUROC than ancestry-discrete modeling at large values of $r$ and~$o$. This is because ancestry-discrete modeling fails to capture the varying omics-endpoint relationship within an ancestry group.

In addition, Figure~\ref{fig:syn_distance} presents the performance of the
three approaches as a function of the distance between the
ancestry-genetic variable~$\mathbf{s}$ and the centroid
$\boldsymbol{\mu} \in
\{\boldsymbol{\mu}_1, \boldsymbol{\mu}_2, \boldsymbol{\mu}_3\}$
of the ancestry group to which $\mathbf{s}$ belongs.
Ancestry-discrete modeling exhibits performance degradation as
the distance from the centroid increases, whereas the two
ancestry-continuum approaches maintain stable performance
across the range of genetic distances. This is consistent with the observation in~\cite{ding2023polygenic}, which quantified performance degradation of polygenic scores as a function of genetic distance using the UK Biobank~\cite{bycroft2018uk}.

\begin{figure*}[t!]
\centering
  \includegraphics[width=\linewidth]{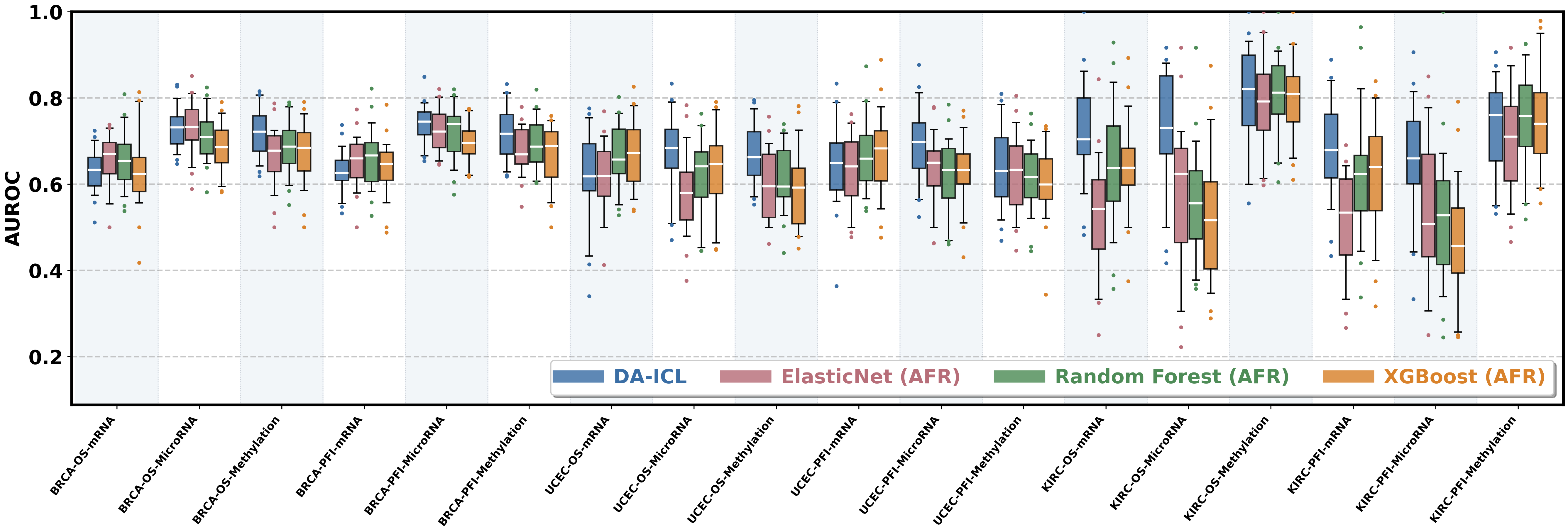}
\caption{Run-level AUROC distributions of DA-ICL for ancestry-continuum modeling and conventional tabular methods for ancestry-discrete modeling across 18 TCGA  tasks, evaluated on AFR test samples. The center line denotes the median, box edges indicate the first and third quartiles, whiskers indicate the non-outlier range, and individual points represent outliers.}
\label{fig:DACE_ICL_and_traditional_methods}
\end{figure*}

\begin{table*}[h!]
\centering
\small
\setlength{\tabcolsep}{2pt}
\renewcommand{\arraystretch}{1.1}
\begin{tabular*}{\textwidth}{@{\extracolsep{\fill}}l*{18}{c}}
\hline
& \rotatebox{60}{\fontsize{5}{6}\selectfont KIRC-OS-mRNA}
& \rotatebox{60}{\fontsize{5}{6}\selectfont KIRC-PFI-mRNA}
& \rotatebox{60}{\fontsize{5}{6}\selectfont UCEC-OS-mRNA}
& \rotatebox{60}{\fontsize{5}{6}\selectfont UCEC-PFI-mRNA}
& \rotatebox{60}{\fontsize{5}{6}\selectfont BRCA-OS-mRNA}
& \rotatebox{60}{\fontsize{5}{6}\selectfont BRCA-PFI-mRNA}
& \rotatebox{60}{\fontsize{5}{6}\selectfont KIRC-OS-miRNA}
& \rotatebox{60}{\fontsize{5}{6}\selectfont KIRC-PFI-miRNA}
& \rotatebox{60}{\fontsize{5}{6}\selectfont UCEC-OS-miRNA}
& \rotatebox{60}{\fontsize{5}{6}\selectfont UCEC-PFI-miRNA}
& \rotatebox{60}{\fontsize{5}{6}\selectfont BRCA-OS-miRNA}
& \rotatebox{60}{\fontsize{5}{6}\selectfont BRCA-PFI-miRNA}
& \rotatebox{60}{\fontsize{5}{6}\selectfont KIRC-OS-Meth}
& \rotatebox{60}{\fontsize{5}{6}\selectfont KIRC-PFI-Meth}
& \rotatebox{60}{\fontsize{5}{6}\selectfont UCEC-OS-Meth}
& \rotatebox{60}{\fontsize{5}{6}\selectfont UCEC-PFI-Meth}
& \rotatebox{60}{\fontsize{5}{6}\selectfont BRCA-OS-Meth}
& \rotatebox{60}{\fontsize{5}{6}\selectfont BRCA-PFI-Meth} \\
\hline
{\fontsize{7}{8}\selectfont DA-ICL vs EN}
& \shortstack{\checkmark\\[2pt]{\fontsize{5}{6}\selectfont$<$\!0.001}}
& \shortstack{\checkmark\\[2pt]{\fontsize{5}{6}\selectfont$<$\!0.001}}
& \shortstack{--\\[2pt]{\fontsize{5}{6}\selectfont 0.829}}
& \shortstack{--\\[2pt]{\fontsize{5}{6}\selectfont 0.461}}
& \shortstack{--\\[2pt]{\fontsize{5}{6}\selectfont 0.059}}
& \shortstack{$\times$\\[2pt]{\fontsize{5}{6}\selectfont 0.034}}
& \shortstack{\checkmark\\[2pt]{\fontsize{5}{6}\selectfont$<$\!0.001}}
& \shortstack{\checkmark\\[2pt]{\fontsize{5}{6}\selectfont$<$\!0.001}}
& \shortstack{\checkmark\\[2pt]{\fontsize{5}{6}\selectfont$<$\!0.001}}
& \shortstack{\checkmark\\[2pt]{\fontsize{5}{6}\selectfont$<$\!0.001}}
& \shortstack{--\\[2pt]{\fontsize{5}{6}\selectfont 0.746}}
& \shortstack{--\\[2pt]{\fontsize{5}{6}\selectfont 0.089}}
& \shortstack{--\\[2pt]{\fontsize{5}{6}\selectfont 0.502}}
& \shortstack{--\\[2pt]{\fontsize{5}{6}\selectfont 0.400}}
& \shortstack{\checkmark\\[2pt]{\fontsize{5}{6}\selectfont$<$\!0.001}}
& \shortstack{--\\[2pt]{\fontsize{5}{6}\selectfont 0.428}}
& \shortstack{\checkmark\\[2pt]{\fontsize{5}{6}\selectfont$<$\!0.001}}
& \shortstack{\checkmark\\[2pt]{\fontsize{5}{6}\selectfont 0.011}} \\[8pt]
{\fontsize{7}{8}\selectfont DA-ICL vs RF}
& \shortstack{\checkmark\\[2pt]{\fontsize{5}{6}\selectfont 0.004}}
& \shortstack{--\\[2pt]{\fontsize{5}{6}\selectfont 0.088}}
& \shortstack{$\times$\\[2pt]{\fontsize{5}{6}\selectfont 0.021}}
& \shortstack{--\\[2pt]{\fontsize{5}{6}\selectfont 0.187}}
& \shortstack{$\times$\\[2pt]{\fontsize{5}{6}\selectfont 0.005}}
& \shortstack{$\times$\\[2pt]{\fontsize{5}{6}\selectfont 0.013}}
& \shortstack{\checkmark\\[2pt]{\fontsize{5}{6}\selectfont$<$\!0.001}}
& \shortstack{\checkmark\\[2pt]{\fontsize{5}{6}\selectfont$<$\!0.001}}
& \shortstack{\checkmark\\[2pt]{\fontsize{5}{6}\selectfont 0.008}}
& \shortstack{\checkmark\\[2pt]{\fontsize{5}{6}\selectfont$<$\!0.001}}
& \shortstack{\checkmark\\[2pt]{\fontsize{5}{6}\selectfont 0.036}}
& \shortstack{--\\[2pt]{\fontsize{5}{6}\selectfont 0.074}}
& \shortstack{--\\[2pt]{\fontsize{5}{6}\selectfont 0.938}}
& \shortstack{--\\[2pt]{\fontsize{5}{6}\selectfont 0.495}}
& \shortstack{\checkmark\\[2pt]{\fontsize{5}{6}\selectfont 0.002}}
& \shortstack{--\\[2pt]{\fontsize{5}{6}\selectfont 0.187}}
& \shortstack{\checkmark\\[2pt]{\fontsize{5}{6}\selectfont 0.001}}
& \shortstack{\checkmark\\[2pt]{\fontsize{5}{6}\selectfont 0.007}} \\[8pt]
{\fontsize{7}{8}\selectfont DA-ICL vs XGB}
& \shortstack{\checkmark\\[2pt]{\fontsize{5}{6}\selectfont 0.001}}
& \shortstack{--\\[2pt]{\fontsize{5}{6}\selectfont 0.063}}
& \shortstack{$\times$\\[2pt]{\fontsize{5}{6}\selectfont 0.026}}
& \shortstack{--\\[2pt]{\fontsize{5}{6}\selectfont 0.335}}
& \shortstack{--\\[2pt]{\fontsize{5}{6}\selectfont 0.877}}
& \shortstack{--\\[2pt]{\fontsize{5}{6}\selectfont 0.332}}
& \shortstack{\checkmark\\[2pt]{\fontsize{5}{6}\selectfont$<$\!0.001}}
& \shortstack{\checkmark\\[2pt]{\fontsize{5}{6}\selectfont$<$\!0.001}}
& \shortstack{\checkmark\\[2pt]{\fontsize{5}{6}\selectfont 0.030}}
& \shortstack{\checkmark\\[2pt]{\fontsize{5}{6}\selectfont$<$\!0.001}}
& \shortstack{\checkmark\\[2pt]{\fontsize{5}{6}\selectfont$<$\!0.001}}
& \shortstack{\checkmark\\[2pt]{\fontsize{5}{6}\selectfont$<$\!0.001}}
& \shortstack{--\\[2pt]{\fontsize{5}{6}\selectfont 0.821}}
& \shortstack{--\\[2pt]{\fontsize{5}{6}\selectfont 0.328}}
& \shortstack{\checkmark\\[2pt]{\fontsize{5}{6}\selectfont$<$\!0.001}}
& \shortstack{--\\[2pt]{\fontsize{5}{6}\selectfont 0.185}}
& \shortstack{\checkmark\\[2pt]{\fontsize{5}{6}\selectfont$<$\!0.001}}
& \shortstack{\checkmark\\[2pt]{\fontsize{5}{6}\selectfont$<$\!0.001}} \\[8pt]
\hline
\end{tabular*}
\caption{Paired $t$-test results comparing DA-ICL with conventional tabular baselines
under ancestry-discrete modeling for the AFR group.
A \checkmark\ indicates that DA-ICL significantly outperforms the baseline
at $p<0.05$;
$\times$ indicates significant underperformance;
-- indicates no significant difference.}
\label{tab:ttest_combined}
\end{table*}
\subsection{Experiments with real-world data}
\label{sec:exp_real}
Conventional ancestry-continuum modeling requires fitting a local model
for every query. DA-ICL instead conditions TabPFN~2.6~\cite{priorlabs_tabpfn26},
a pretrained ICL model~\cite{hollmann2025accurate}, on a query-specific
context. All ICL experiments use 32 estimators and softmax temperature
0.9, fixed across tasks and runs.

For each of 30 seeds, patients are randomly divided into three folds;
fold~0 is held out and the other two form the development pool. We report
performance for AFR patients in fold~0. Conventional baselines use an
internal development subset for hyperparameter selection, whereas the
TabPFN settings are not selected using validation outcomes.

ICL-Discrete uses all AFR development patients. ICL-Continuum uses the
$K=100$ nearest patients from the combined White and AFR development
pool in Euclidean genotype-PCA30 space. DA-ICL-Continuum starts from
the same KNN100 context and searches its complement in AE20 space.
Within each outcome class, a separate MST is constructed and candidates
incident to edges crossing into the KNN context are added. Labels define
the strata but are not distance coordinates. TabPFN then uses the selected
patients' omics-PCA100 features and labels to predict the query's
omics-PCA100 features. Test patients are excluded from contexts and candidate pools; their
outcomes are used only for evaluation, never for neighbor selection,
DACE, fitting, or tuning. Baselines comprise ancestry-discrete elastic
net, XGBoost, and random forest, plus ICL without DACE under discrete
and continuum settings.

We first compare ancestry-continuum modeling using DA-ICL, evaluated on samples from the African~(AFR) group, against random forest, elastic net, and XGBoost models developed under the ancestry-discrete modeling setting. Figure~\ref{fig:DACE_ICL_and_traditional_methods} shows that DA-ICL achieves the highest median AUROC on 11 of the 18 tabular omics-based tasks from TCGA when evaluated on AFR samples. In addition, we conduct paired $t$-tests between continuum modeling using DA-ICL and ancestry-discrete modeling using the three conventional tabular methods. Table~\ref{tab:ttest_combined} indicates that DA-ICL achieves significantly better results in at least 9 tasks, although the results vary by task and baseline.

Next, we compare ICL without DACE under ancestry-discrete modeling, ICL without DACE under ancestry-continuum modeling, and DA-ICL under ancestry-continuum modeling. Table~\ref{tab:auroc_comparison} shows that DA-ICL under ancestry-continuum modeling achieves the highest mean AUROC on 10 out of 18 tasks, while ICL without DACE under ancestry-discrete modeling achieves the highest mean AUROC on 6 tasks. ICL without DACE under ancestry-continuum modeling achieves the highest mean AUROC on 2 of the 18 tasks, suggesting that a neighborhood-only context may be insufficient for some tasks. These results indicate that, for some real-world omics-based prediction tasks, constructing contexts using only samples near each target ancestry point in the genetic space may be insufficient to characterize the relationship between omics features and endpoint labels. DACE enriches the local context and yields the highest mean AUROC in a majority, but not all, of the evaluated tasks.
\begin{table*}[t!]
\centering
\small
\setlength{\tabcolsep}{3pt}
\renewcommand{\arraystretch}{1.15}

\resizebox{\textwidth}{!}{%
\begin{tabular}{l*{18}{c}}
\hline
& \rotatebox{50}{KIRC-OS-mRNA}
& \rotatebox{50}{KIRC-PFI-mRNA}
& \rotatebox{50}{UCEC-OS-mRNA}
& \rotatebox{50}{UCEC-PFI-mRNA}
& \rotatebox{50}{BRCA-OS-mRNA}
& \rotatebox{50}{BRCA-PFI-mRNA}
& \rotatebox{50}{KIRC-OS-miRNA}
& \rotatebox{50}{KIRC-PFI-miRNA}
& \rotatebox{50}{UCEC-OS-miRNA}
& \rotatebox{50}{UCEC-PFI-miRNA}
& \rotatebox{50}{BRCA-OS-miRNA}
& \rotatebox{50}{BRCA-PFI-miRNA}
& \rotatebox{50}{KIRC-OS-Meth}
& \rotatebox{50}{KIRC-PFI-Meth}
& \rotatebox{50}{UCEC-OS-Meth}
& \rotatebox{50}{UCEC-PFI-Meth}
& \rotatebox{50}{BRCA-OS-Meth}
& \rotatebox{50}{BRCA-PFI-Meth} \\
\hline

ICL-Discrete
& $0.68\pm0.09$
& $0.66\pm0.10$
& $\mathbf{0.69}\pm0.08$
& $\mathbf{0.68}\pm0.09$
& $\mathbf{0.67}\pm0.05$
& $0.66\pm0.05$
& $0.59\pm0.13$
& $0.51\pm0.15$
& $0.63\pm0.08$
& $0.64\pm0.09$
& $0.73\pm0.05$
& $0.73\pm0.05$
& $\mathbf{0.82}\pm0.08$
& $\mathbf{0.76}\pm0.10$
& $0.64\pm0.07$
& $\mathbf{0.64}\pm0.08$
& $0.71\pm0.06$
& $0.69\pm0.06$ \\

ICL-Continuum
& $0.71\pm0.13$
& $0.61\pm0.15$
& $0.65\pm0.09$
& $0.66\pm0.09$
& $0.66\pm0.05$
& $\mathbf{0.67}\pm0.05$
& $0.70\pm0.14$
& $0.61\pm0.19$
& $0.66\pm0.09$
& $\mathbf{0.70}\pm0.08$
& $0.70\pm0.06$
& $0.71\pm0.05$
& $0.71\pm0.13$
& $0.61\pm0.10$
& $0.59\pm0.08$
& $0.59\pm0.07$
& $0.70\pm0.06$
& $0.70\pm0.07$ \\

DA-ICL-Cont.
& $\mathbf{0.73}\pm0.11$
& $\mathbf{0.68}\pm0.11$
& $0.62\pm0.11$
& $0.65\pm0.09$
& $0.63\pm0.05$
& $0.63\pm0.05$
& $\mathbf{0.73}\pm0.13$
& $\mathbf{0.66}\pm0.13$
& $\mathbf{0.67}\pm0.09$
& $0.69\pm0.09$
& $\mathbf{0.73}\pm0.05$
& $\mathbf{0.74}\pm0.05$
& $0.80\pm0.12$
& $0.73\pm0.11$
& $\mathbf{0.67}\pm0.07$
& $0.64\pm0.10$
& $\mathbf{0.72}\pm0.06$
& $\mathbf{0.72}\pm0.06$ \\
\hline
\end{tabular}%
}
\caption{AUROC (mean $\pm$ std) over 30 runs on 18 TCGA tasks for AFR test patients. ICL-Discrete uses all AFR development patients; ICL-Continuum uses query-specific KNN100 contexts from White and AFR development patients; DA-ICL-Cont. adds DACE-selected patients to these contexts. Bold indicates the highest mean AUROC per task.}
\label{tab:auroc_comparison}
\end{table*}
\begin{figure*}[h]
\centering
\includegraphics[width=0.327\linewidth]{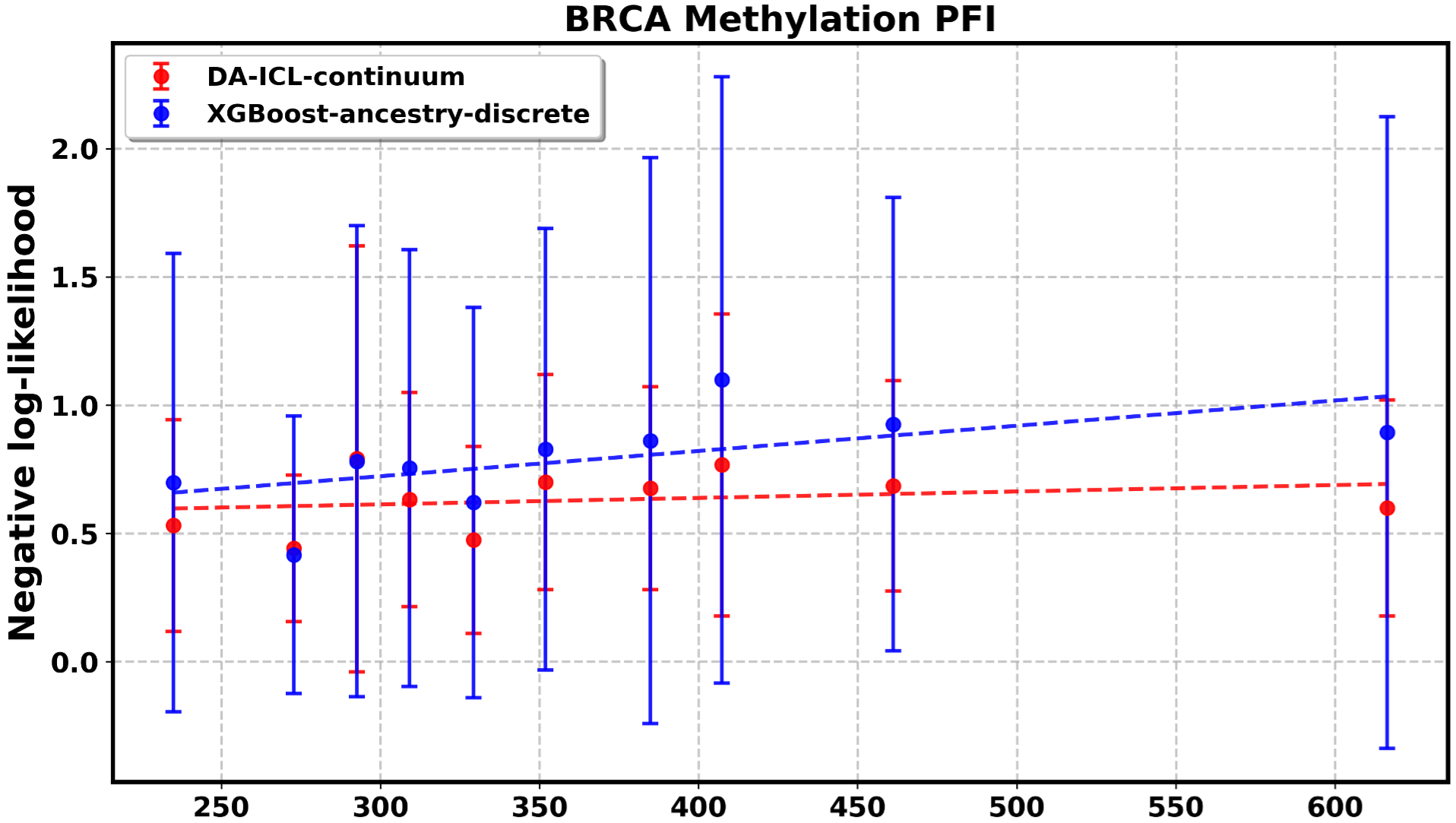}
\includegraphics[width=0.315\linewidth]{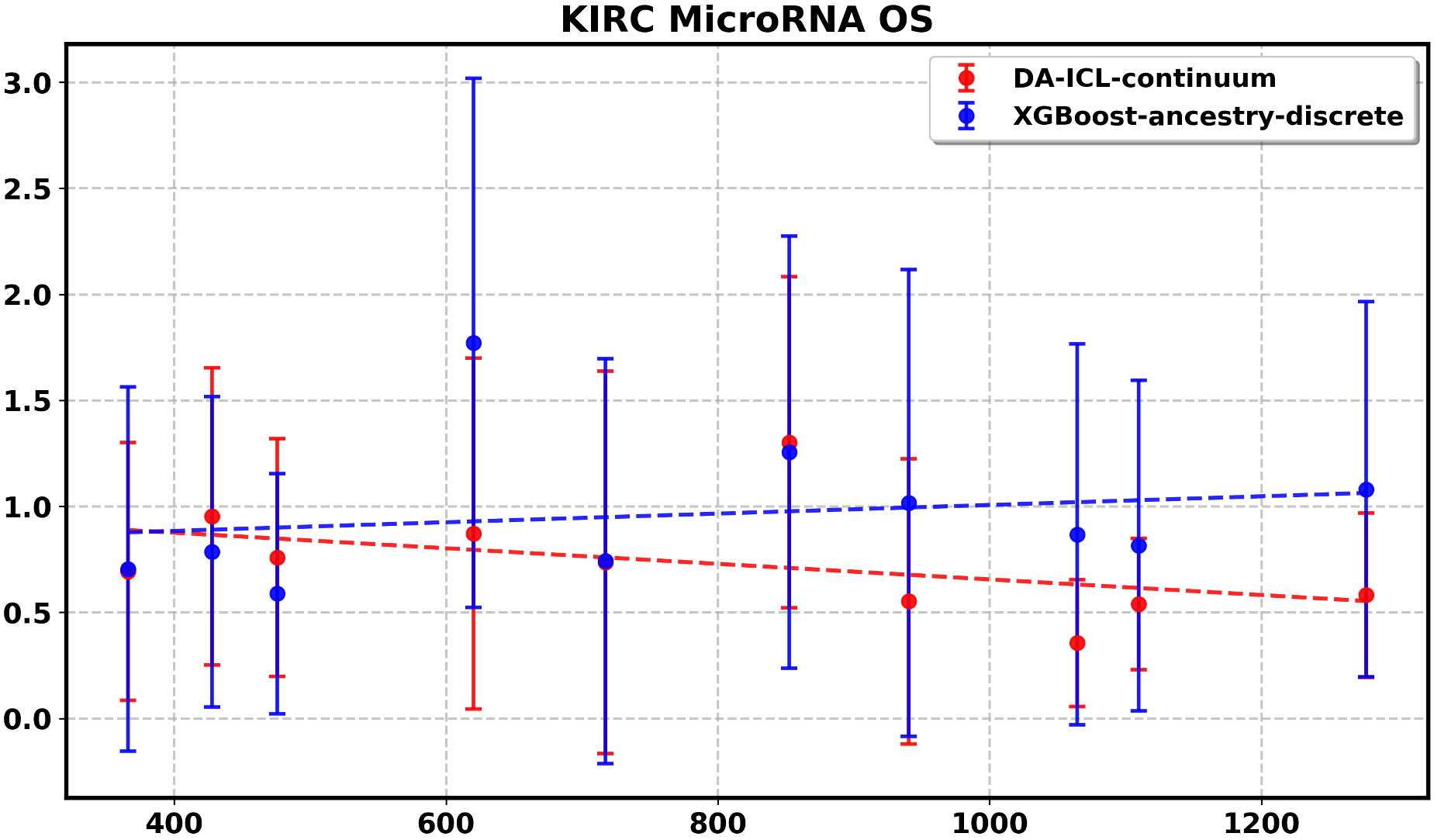}
\includegraphics[width=0.315\linewidth]{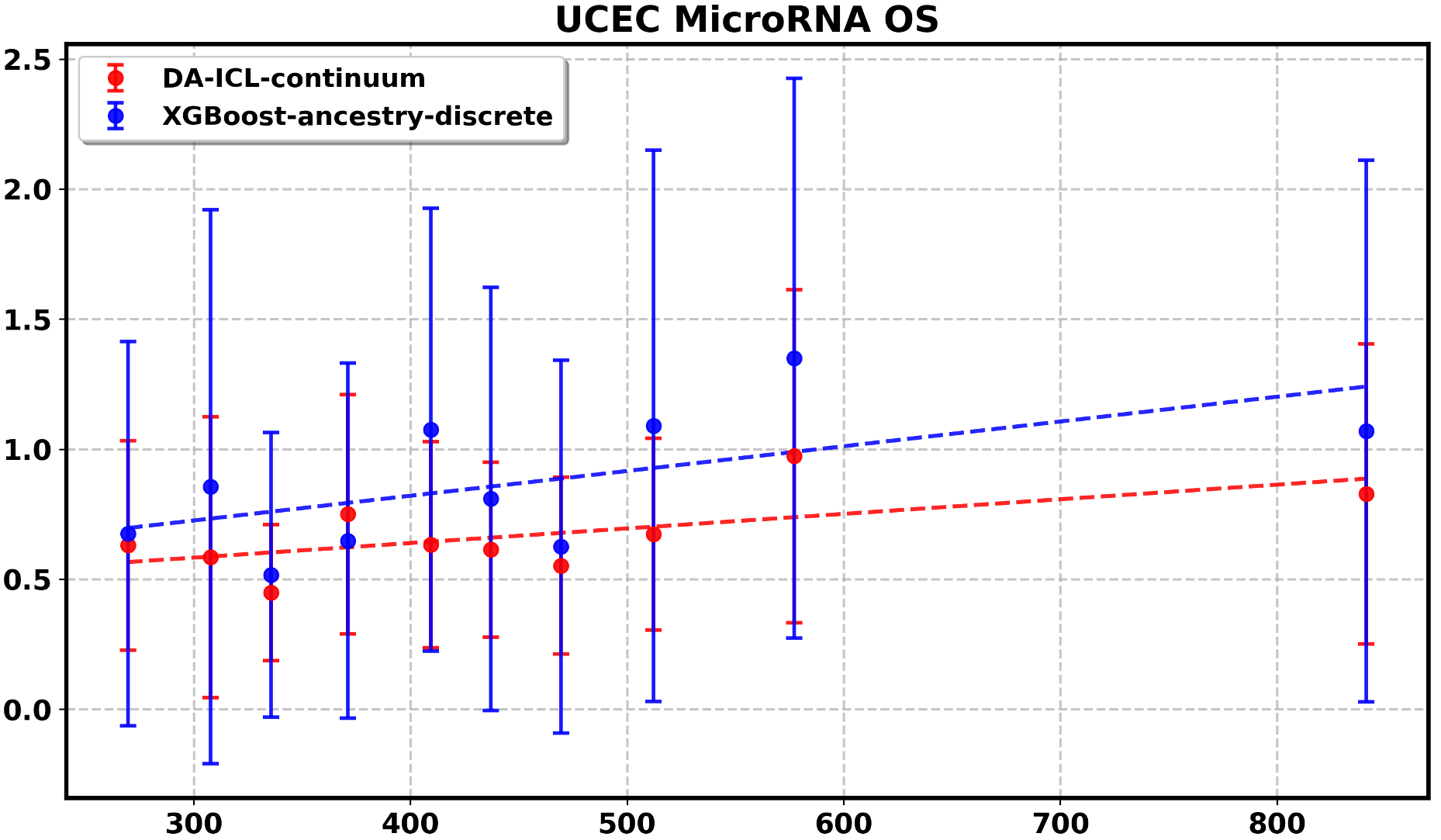}\\
\includegraphics[width=0.327\linewidth]{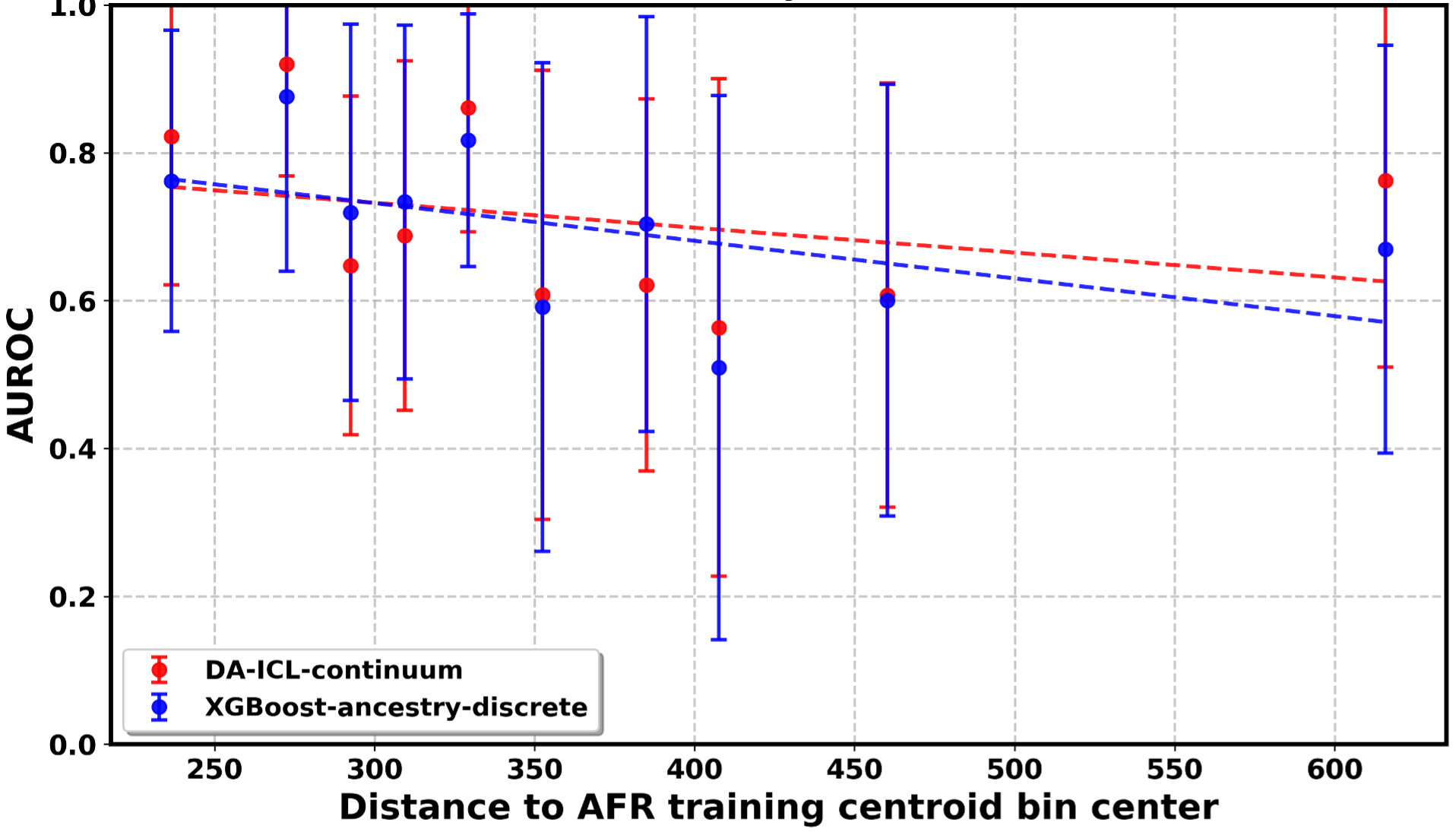}
\includegraphics[width=0.315\linewidth]{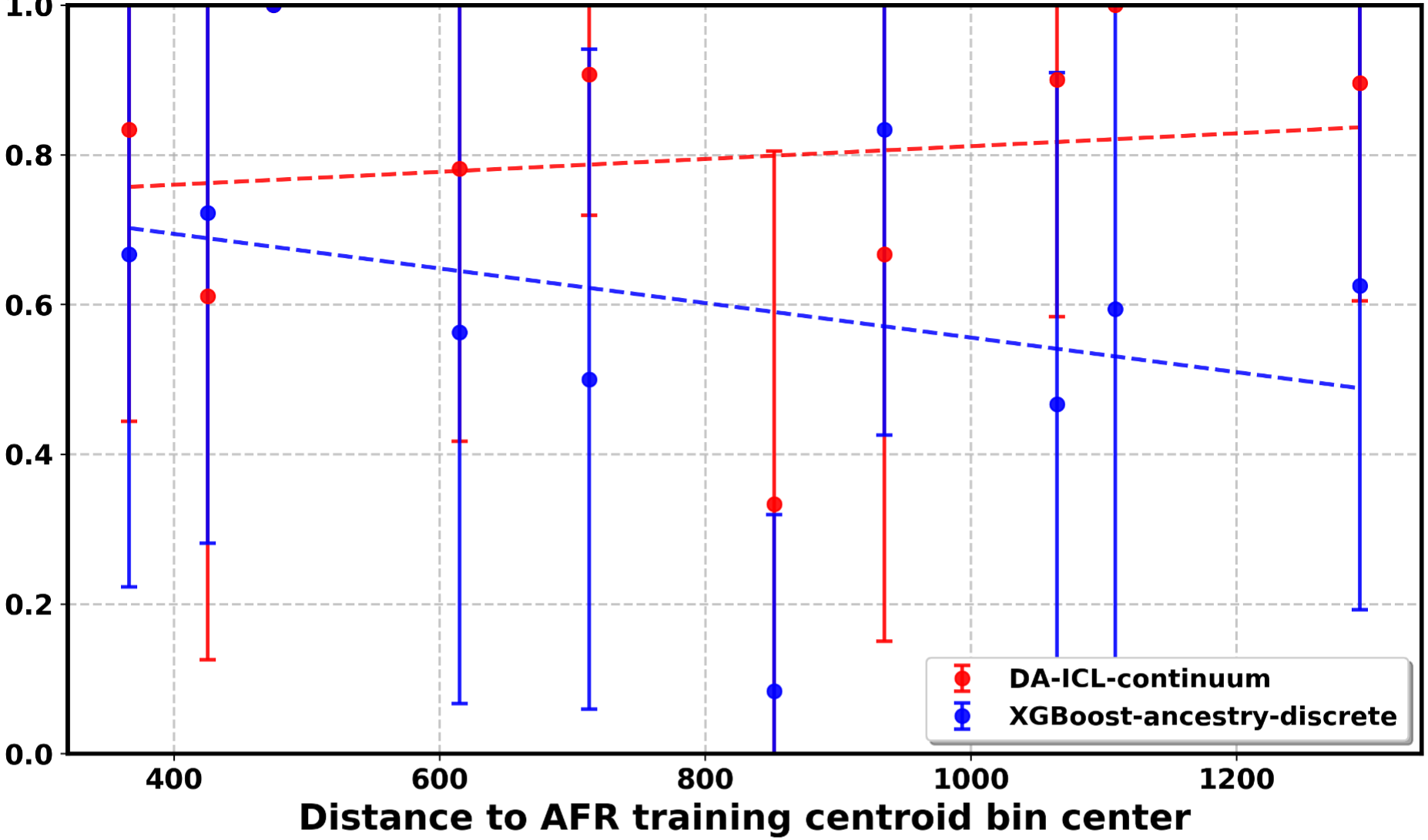}
\includegraphics[width=0.315\linewidth]{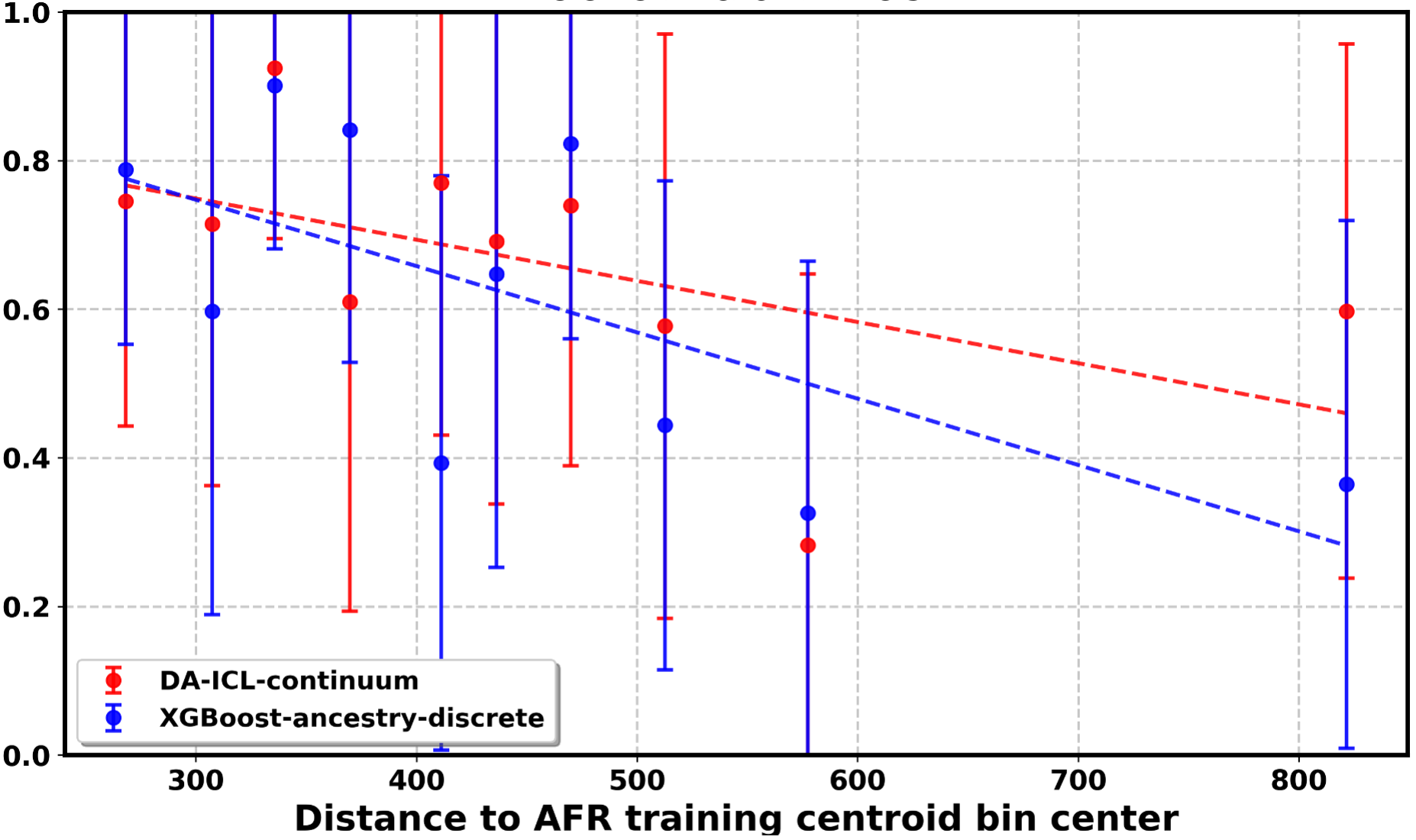}
\caption{Negative log-likelihoods (NLLs; top) and AUROCs (bottom) of DA-ICL under the ancestry-continuum setting and XGBoost under the ancestry-discrete setting. DA-ICL maintains relatively stable performance with increasing genetic distance, whereas XGBoost degrades more rapidly, with increasing NLL and decreasing AUROC. 
}
\label{fig:real_genetic_distance}
\end{figure*}

Lastly, we compare XGBoost under the ancestry-discrete setting against DA-ICL under the ancestry-continuum setting as a function of genetic distance to the centroid of the AFR samples, using test samples from the AFR group. We divide the genetic distances from the test samples to the AFR centroid into ten bins, and evaluate AUROC and NLL within each bin across 30 runs. Figure~\ref{fig:real_genetic_distance} shows that DA-ICL under ancestry-continuum modeling exhibits a slower trend of performance degradation with increasing genetic distance for both NLL and AUROC.

\section{Conclusion}
We present DA-ICL for ancestry-continuum modeling. We show theoretically that the Bayes-optimal ancestry-continuum classifier has error no greater than the Bayes-optimal ancestry-discrete classifier under the stated data-generating process. DA-ICL enables omics-based prediction at each ancestral point without target-task parameter updates to the foundation model and uses distribution-aware context enhancement to enrich local contexts. In the TCGA evaluation, DA-ICL achieves the highest mean AUROC in 10 of 18 tasks, indicating task-dependent empirical benefits while preserving the efficiency of in-context learning.

\section*{Acknowledgment}
This work was supported by the US National Cancer Institute grant R01CA262296.
\bibliographystyle{abbrv}
\bibliography{Reference}

\end{document}